\documentclass[sn-mathphys]{sn-jnl}% Math and Physical Sciences Reference Style

\jyear{2022}%

\theoremstyle{thmstyleone}%
\theoremstyle{thmstyletwo}%
\usepackage[T1]{fontenc}
\theoremstyle{thmstylethree}%
\usepackage{amssymb}
\usepackage{amsfonts}
\usepackage{amsmath}
\usepackage{algorithm}
\usepackage{algpseudocode}
\begin{document}
\title[UW-CVGAN]{UW-CVGAN: UnderWater Image Enhancement with Capsules Vectors Quantization}

%%=============================================================%%
%% Prefix	-> \pfx{Dr}
%% GivenName	-> \fnm{Joergen W.}
%% Particle	-> \spfx{van der} -> surname prefix
%% FamilyName	-> \sur{Ploeg}
%% Suffix	-> \sfx{IV}
%% NatureName	-> \tanm{Poet Laureate} -> Title after name
%% Degrees	-> \dgr{MSc, PhD}
%% \author*[1,2]{\pfx{Dr} \fnm{Joergen W.} \spfx{van der} \sur{Ploeg} \sfx{IV} \tanm{Poet Laureate} 
%%                 \dgr{MSc, PhD}}\email{iauthor@gmail.com}
%%=============================================================%%

\author*[1]{\fnm{Rita} \sur{Pucci}}\email{rita.pucci@uniud.it}
% \equalcont{These authors contributed equally to this work.}
\author[1]{\fnm{Christian} \sur{Micheloni}}\email{christian.micheloni@uniud.it}

\author[1]{\fnm{Niki} \sur{Martinel}}\email{niki.martinel@uniud.it}
% \equalcont{These authors contributed equally to this work.}

\affil[1]{\orgdiv{Department of computer science}, \orgname{University of Udine}, \orgaddress{\street{via delle scienze}, \city{Udine}, \state{Italy}}}

\def\myAE{\texttt{AE}}
\def\myVAE{\texttt{VAE}}
\def\myCVGAN{\texttt{CV-GAN}}
\def\myUWCVGAN{\texttt{UW-CVGAN}}
\def\myUWVQGAN{\texttt{UW-VQGAN}}
\def\myVQGAN{\texttt{VQ-GAN}}
\def\myVQVAE{\texttt{VQ-VAE}}
\def\myGAN{\texttt{GAN}}
\def\myRbA{\texttt{RbA}}
%%==================================%%
%% sample for unstructured abstract %%
%%==================================%%

\abstract{The degradation in the underwater images is due to wavelength-dependent light attenuation, scattering, and to the diversity of the water types in which they are captured. Deep neural networks take a step in this field, providing autonomous models able to achieve the enhancement of underwater images. We introduce Underwater Capsules Vectors GAN \myUWCVGAN{ }based on the discrete features quantization paradigm from \myVQGAN{ }for this task. The proposed \myUWCVGAN{ }combines an encoding network, which compresses the image into its latent representation, with a decoding network, able to reconstruct the enhancement of the image from the only latent representation. In contrast with \myVQGAN, \myUWCVGAN{ }achieves the features quantization by exploiting the clusterization ability of capsules layer, making the model completely trainable and easier to manage. The model obtains enhanced underwater images with high quality and fine details. Moreover, the trained encoder is independent from the decoder giving the possibility to be embedded onto the collector as compressing algorithm to reduce the memory space required for the images, of factor $3\times$. \myUWCVGAN{ }is validated with quantitative and qualitative analysis on benchmark datasets, and we present metrics results compared with the state of the art.}

% Open Source Flickr images dataset labelled ”underwater” and
%%================================%%
%% Sample for structured abstract %%
%%================================%%


%%\pacs[JEL Classification]{D8, H51}

%%\pacs[MSC Classification]{35A01, 65L10, 65L12, 65L20, 65L70}

\maketitle

\section{Introduction}\label{sec1}
The Oceans cover most of the planet where we live and are as fascinating as harsh, complex, and dangerous for exploration. For these reasons, in recent years, the exploration and protection of the rich ecosystems result in sophisticated visual sensing systems used to capture information from the underwater world. These systems are often embedded in robots that are an attractive option because of their non-intrusive, passive, and energy-efficient nature. This underwater tools are efficient for monitoring the coral barrier reef \cite{shkurti2012multi}, exploring the depth of the ocean \cite{whitcomb2000advances}, and analysing the seabed \cite{bingham2010robotic}. However, such methods are affected by the influence of water for light absorption and scattering that cause visible alteration in colours and definition of the images quality. As a consequence, underwater images often are hazy and show greenish or bluish tinged. For this reason, an intriguing challenge in computer vision is the enhancement task of these data.

\begin{figure}
  \centering
  \includegraphics[width=.8\linewidth]{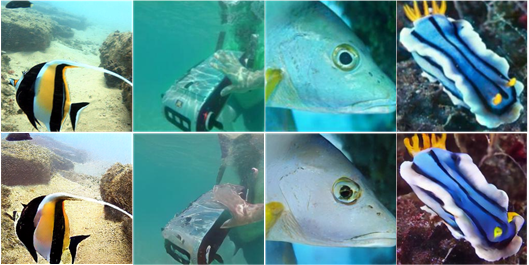}
  \caption{The first row shows the effects of water light refraction in underwater images. The second row shows the results of the proposed approach in removing the blurriness and cold tones due to the colours wavelength suppression.}
  \label{fig:Figure1}
\end{figure}
Existing methods for underwater images enhancement include both traditional techniques (not based on machine learning algorithms) and deep learning-based techniques. The traditional techniques consist of non physical based methods, which adjust image pixel values to obtain visually neat images~\cite{li2016underwater,ghani2015underwater}, and physical based methods, which provide a precise restoration of the images requiring various complex underwater physical and optical factors~\cite{han2017active, neumann2018fast}. The former apply optical functions that are not generalised among different datasets, and the latter techniques need abundant data in order to extract the necessary factors~\cite{hu2021underwater}. 
In contrast, deep learning-based techniques are generalised over the data and have been recently applied for underwater image enhancement~\cite{islam2020fast,park2019adaptive,hu2021underwater,fabbri2018enhancing,zhang2021dugan,zhu2017unpaired,islam2020simultaneous,guo2019underwater} with good results. These techniques automatically extract information from the images learning how to provide the enhancement of them. 

We focus on deep learning-based techniques and in particular on the Generative Adversarial Networks (\myGAN)~\cite{goodfellow2020generative}, that can automatically discover and learn the regularities or patterns in input data so that the model can be used to generate and reconstruct samples~\cite{yuille2021deep}. In the field of enhancement of underwater images, the main disadvantage by using \myGAN{ }is that the category attribute of the image cannot be controlled, resulting in blurry images. Works at SOTA, mitigate this problem by using conditional information. In~\cite{islam2020fast}, authors perform a fast enhancement of underwater images by adding to the input random noise and associating to the loss function additional aspects rather than just the adversarial one. Other works~\cite{park2019adaptive,hu2021underwater,zhang2021dugan} are based on a cycle-consistent adversarial network (CycleGAN)~\cite{zhu2017unpaired} that demonstrate to obtain good results. They require a multi-term objective function for the generator loss, to correct colour casts effectively, and to improve image quality. In~\cite{islam2020simultaneous}, the authors provide a simultaneous enhancement and super-resolution of the underwater image by a fully convolutional encoder-decoder architecture. In contrast with these works, we want to avoid the conditioning of the model by using only the underwater images without additional information. This reduces the amount of data that has to be set for the enhancement process.

Further works avoid the conditioning by using U-Net shape to keep spatial information through the reconstructive process~\cite{guo2019underwater,islam2020fast, fabbri2018enhancing}. In~\cite{guo2019underwater}, the authors proposed a multiscale dense block to concatenate the features among the features extraction and features reconstruction phases. In ~\cite{islam2020fast, fabbri2018enhancing}, the authors input only the underwater image and use a U-Net based structure to enhance it using skip connections to conserve the spatial information among the reconstruction layers. In fact, models based on U-Net structure create a strong connection between the features extraction and the image reconstruction phases, that do not give the possibility to separate the two phases and to obtain a compression algorithm. We think that the compression of the images avoid wasting memory space for useless information and for this idea, we propose a model that provides the compression of the images to their features representations and that is able to reconstruct the enhanced image from them. 

For such a model, we consider the  Variational AutoEncoder (\myVAE) family models. In fact, the \myVAE{ }consists of encoder-decoder networks which learn to represent the input data in a latent representation and reconstruct them in the output. The latent representation is extracted through encoder by the latent space, which describes the probability distribution of the data learned by the model in the training phase. It is worth noting that in \myVAE, the encoder and the decoder are two separated networks. 

The Variational Quantized AutoEncoder\myVQVAE{ }~\cite{oord2017neural} encodes features as discrete latent variables by a quantized latent space and offers a direct control over the information content of the learned representation. Even if the quantization provides a good property of features separation~\cite{tjandra2019vqvae,eloff2019unsupervised}, a discrete latent representation is not differentiable, making the training phase challenging. There exist strategies to circumvent the differentiability problem, such as optimise the mean output of discrete random units~\cite{jang2016categorical}, altering the training dynamics methods~\cite{lancucki2020robust}, or applying a straight-through gradient estimator, which copies the gradients from the decoder to the encoder~\cite{esser2021taming,oord2017neural}. We think that the idea of quantization of the latent space used for features representation is a good strategy to obtain a model able to reconstruct high quality images and that can be applied to extract the fundamental features from underwater images leaving the noise of water behind. However the non differentibility represents a limit to the possibilities of the model, and we propose a model that takes into consideration this aspect.

We propose \myCVGAN, a generative \myVAE{ }model based on Capsules Vectors~\cite{sabour2017dynamic,pucci2021fixed,pucci2020deep,deng2018hyperspectral} that exploits the quantization idea by maintaining the model completely differentiable. Works in image analysis~\cite{pucci2021collaborative,pucci2022pro} exploit capsules for colours reconstruction and image classification and demonstrate the capsules' clusterization ability.

\myCVGAN{ } is a tradeoff between a fully differentiable \myVAE{ }and a discrete latent representation. Our model provides a learned continuous latent space where the features quantization is implemented by capsules layer. This layer represents each input image as the clustered probability distribution of features which represent the entities in the image. In particular, we replace the discrete quantized vectors proposed in \myVQVAE{ }with the capsules vectors by replacing the code-book used in \myVQVAE{ }with the capsule layer. The clusterization of these vectors is obtained by the routing-by-agreement (\myRbA)~\cite{sabour2017dynamic}. With \myRbA{}, the layer learns a dynamic representation of the features clustering them into the capsules vectors. 
We demonstrate that the capsules vectors is the latent representation of the image and this representation can be used for the reconstruction of the enhancement of the image. 

In contrast to~\cite{islam2020fast, fabbri2018enhancing}, our model needs only the latent compression of the image for the reconstruction of the enhancement of the image, and the encoder network can be used as compression algorithm for the image collected. The compression obtained is of $3\times$ of the original dimension of the images; we identify~\cite{islam2020simultaneous} which provide good metric results but with a lower compresses of the original dimension. The use of \myVAE{ }models for the enhancement task is presented in~\cite{mello2020alternative, kingma2013auto}, with not generative models conditioned by the simulation of the effects of optical factors. 
 
The application of \myCVGAN{ }model for the task of underwater images enhancement is here after denoted by \myUWCVGAN. \myUWCVGAN{ }outperforms the results on metrics obtained with \myVQGAN in the proposed task and both following the training phase described in Sec.~\ref{dataset}. Moreover, our model is differentiable, still maintaining the features quantization characteristic. We evaluate \myUWCVGAN{ }on benchmarks datasets and we compare the results with \myGAN{ }models proposed for enhancement of underwater images.

The main contributions of this paper are summarised as follows.: 
\begin{itemize}
    \item we propose a novel \myVAE{ } model based on capsules layer named \myUWCVGAN{ } which exploits the \myRbA{ } procedure as clusterization paradigm for the enhancement of underwater images; 
    \item we demonstrate that the new model is completely differentiable, still implementing the quantization of the latent space;
    \item we demonstrate that the encoder provides a latent representation of the image that is sufficient to the decoder to reconstructs the image enhanced in quality;
    \item the separated encoder-decoder networks in \myUWCVGAN{ }give the possibility to embed the encoder network into the collector to compress the images by a factor of $3\times$ storing only the latent representation. 
\end{itemize}
This is the first attempt of applying capsules to learn a latent space and, in particular, for image enhancement. We think that the clusterization offered by the capsules layer is a valid candidate for the latent space implementation and that the results obtained in underwater image enhancement demonstrate the ability of the model in features extraction. A detailed discussion will be presented in Sec.~\ref{C_Q}.

\section{Background} 
% In this work, we propose a bottleneck for \myVAE~\cite{kingma2013auto,rezende2014stochastic} based on capsules layers~\cite{sabour2017dynamic} which provides continuous latent variables~\cite{mnih2014neural} to enhance underwater images. The proposed bottleneck learns to cluster useful features from the input image. We briefly introduce the basic concept embraced in our idea. 
\subsection{Underwater images}
\begin{figure}
  \centering
  \includegraphics[width=.8\linewidth]{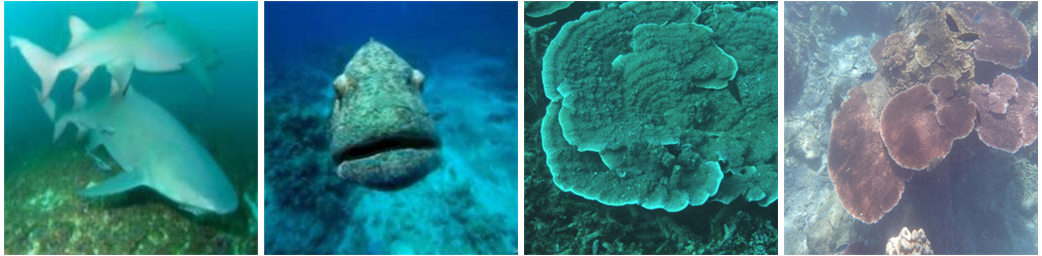}
  \caption{Image samples shot in underwater environment. These samples show some of the typical problem with underwater images, from the left: low edge definition, blueish colours, greenish colours, blurriness. These images are from EUVP,UFO120,HICRD,UIEB datasets.}
  \label{fig:UWIN}
\end{figure}

Underwater images are characterised by a whole range of light distortions due to the water absorption of light waves.  The colours and the edges' perception change with depth, illumination, and turbidity of water making the image appear completely different. The low definition of edges, the distortion of colours in a blueish or greenish colourisation, and the distortion of light make the identification of subjects in the image difficult.
In Fig.~\ref{fig:UWIN}, we show some samples of the high variability in visual scenes that may occur in underwater environments. 
It is necessary for the vision-based algorithms in this field, to reach a good generalisation in order to work within this range of distortions. Some physics-based models in this field consider the Jaffe-McGlamery model~\cite{jaffe1990computer,mcglamery1975computer} which is able to describe the underwater image degradation  by defining the distorted images as follows:
\begin{equation}
   I(x) = J(x)t(x) + \alpha(1 - t(x)) 
\end{equation}
where $I(x)$ denotes the degraded image, and $J(x)$ is the clear image. The parameter $\alpha$ is the global atmospheric light and it indicates the intensity of ambient light, and the $t (x) \in [0, 1]$ is the transmission map matrix which denotes the percentage of the scene radiance reaching the camera. It is defined as: 
\begin{equation}
 t (x) = e^{-\nu d(x)} 
\end{equation}
where $\nu$ is the atmospheric attenuation coefficient and $d(x)$ is the distance of the object to camera. 
Even if this model is straightly related to the real nature of the underwater distortions in the images, the application of such a model to represent the image is a challenging task due to the high parametrization which requires estimating $\alpha$, $\nu$ and $d(x)$ in different underwater environments. It is to consider that the disparity of the underwater conditions present in nature is often much more complicated to represent than the possibilities offered by this model.
Deep neural networks demonstrate to be a valid alternative to the physics-based models. These models are trained to extract representative features for images without any environmental information, providing good results with a wide range of underwater scenes. 
% In the following section, we focus on the Variational AutoEncoder~\myVAE{ }models that are here proposed for the enhancement of images.  

\subsection{Variational Auto-Encoder (\myVAE)}
\myVAE{ }is a deep convolutional neural network algorithm introduced in~\cite{kingma2013auto}, it belongs to the families of probabilistic graphical models and variational Bayesian methods. This model consists of an encoder ($E$) and a decoder ($G$) networks. The encoder network parameterises a posterior distribution $q(z\vert\textbf{X})$ of latent random variables $z$ given the input data $\textbf{X}$, and a prior distribution $p(z)$. The encoder must learn an efficient compression of the data into this lower-dimensional space, the latent space. The decoder network receives the representation $z$ and outputs the probability distribution of data, so it is denoted by a distribution $p(\textbf{X}\vert z)$. The posteriors and priors in \myVAE{ }are assumed to be normally distributed with diagonal covariance.
\begin{figure}
  \centering
  \includegraphics[width=.8\linewidth]{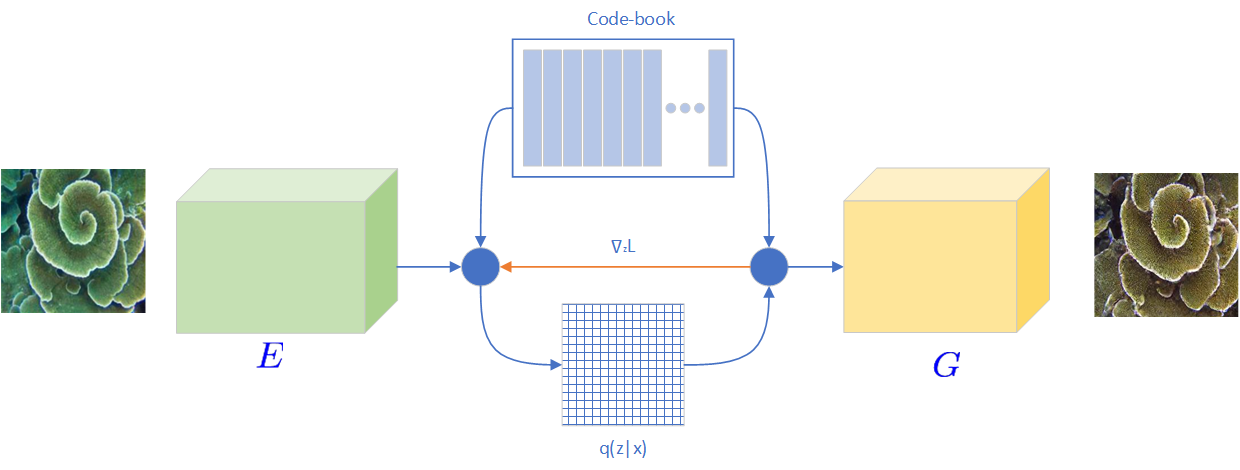}
  \caption{The \myVQVAE~\cite{oord2017neural,esser2021taming} model. The output of the encoder $z^{\textbf{e}(\textbf{X})}$ is quantized based on a code-book randomly initialised. The application of a discrete quantization makes the connections with $q(  .)$ not differentiable. By the straight through estimator, the gradient $\bigtriangledown_{z}L$  will push the encoder to change its output without taking into consideration the quantization step. }
  \label{fig:VQVAE}
\end{figure}
\subsection{Vector Quantized Variational Auto-Encoders (\myVQVAE)}
An interesting architecture based on \myVAE{ }is the \myVQVAE{ }model~\cite{oord2017neural}. These models propose a new latent space where a quantization function is used to represent the latent vectors. The latent space is defined as $\textbf{e}\in \mathbb{R}^{K\times M}$ where $K$ is the size of the discrete latent space (i.e., K-way categories for the quantization), and $M$ is the dimensionality of each latent vector $\textbf{e}^i \in \mathbb{R}^M$, $i \in 1, 2, ..., K$ that compose the code-book, as shown in Fig.~\ref{fig:VQVAE}. The input to $G$ is the corresponding embedding vector $\textbf{e}^k$.

These models are regular autoencoders with a particular non-linearity that maps the latents to 1-of-K code-book. The complete set of parameters for the model is the union of parameters of $E$, $G$, and the embedding space $\textbf{e}$. In contrast with $E$ and $G$ parameters, the embedding space is discrete and is not trainable with the canonical strategy of backpropagation of gradients. This characteristic makes the latent space on one side easier to manage and on the other side difficult to train.
\subsection{Capsules Layer}
A capsule is a set of neurons which capture the presence of an entity in an image~\cite{sabour2017dynamic} by clustering the features extracted. The capsules collectively produce an activation vector with one element from each neuron to hold that neuron's instantiation value. In a hierarchical structure of capsules layers, the activation vector of each capsule of the higher layer makes predictions for the parent capsules. Capsules layers are trained by the \myRbA{ }mechanism. This mechanism compares the higher layer prediction with the activation vector of the parents-capsules. The activation vectors which match are clusterised to provide the final vector that represents the probability of the entities to be present. In Sec.~\ref{C_Q} we demonstrate that the capsules and \myRbA{ }are promising ideas for a dynamic clusterization of the features in \myVQVAE{}.
Such a solution is still not explored at the SOTA in \myVAE{ }and we think that can be a good tradeoff between a continuous latent space and the discrete latent space of \myVQVAE{ }made by a fixed code-book. We propose a capsules layer to cluster the representative variables to compute the distribution of the features as a replacement for the quantization. This makes the training procedure completely differentiable while maintaining the clustering idea of the feature representation.

\subsection{Generative Adversarial Network}
The Generative Adversarial Network~\cite{goodfellow2020generative} is a framework for generative modelling of data through learning a transformation from points belonging to a prior distribution ($z \sim p_z$) to points from a data distribution ($\textbf{X} \sim p_{data}$). It consists of two models that play an adversarial game: a generator $G$ and a discriminator $D$. While $G$ attempts to learn the aforementioned transformation $G(z)$, $D$ acts as a critic $D(\cdot)$ determining whether the sample provided to it is from the generator’s output distribution ($G(z) \sim p_{G}$) or from the data distribution ($\textbf{X} \sim p_{data}$), thus giving a scalar output ($y \in \{0, 1\}$).
The generator wants to fool the discriminator by generating samples that resemble those from the data distribution, while the discriminator wants to accurately distinguish between real and generated data. The two models are neural networks and they play an adversarial game with the objective:
\begin{equation}
    min_{G}max_{D}V(D,G) = E_{\textbf{X} \sim p_{data}(\textbf{X})}[log D(\textbf{X})] + E_{z \sim p_{z}(z)}[log(1 - D(G(z)))] 
\end{equation}

\section{Methods}\label{sec11}
\subsection{Capsules layer architecture}
\label{C_L}
\begin{figure}
  \centering
  \includegraphics[width=.9\linewidth]{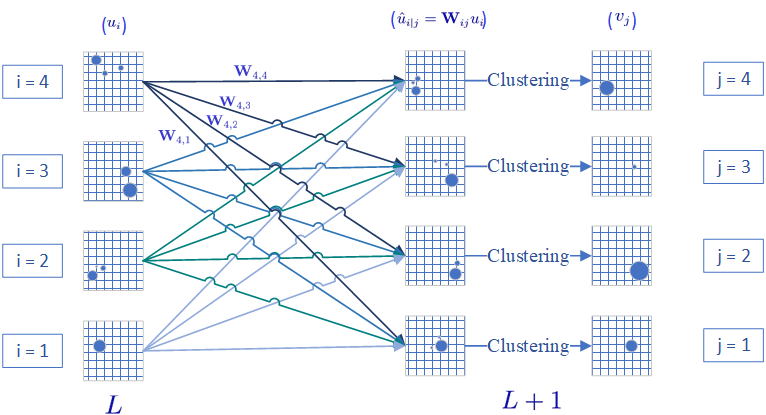}
  \caption{Visual representation of the \myRbA{ }paradigm: the output of the capsules at layer $L$, $\textbf{u}_i$, is forwarded to the capsules at the upper layer $L+1$. An affine transformation is applied to obtain $\hat{\textbf{u}}_{i}$. Then the features are clustered in $\textbf{v}_j$ based on the agreement among capsules.}
  \label{fig:RoutingBAgreement}
\end{figure}

Let $\textbf{u}_i\in \mathbb{R}^{d_u}$ be an output of a capsule $i$ at layer $L$, and $j$ the index at layer $L+1$. The affine transformation of $\textbf{u}_i$ is calculated as:
\begin{equation}
    \hat{\textbf{u}}_{i\vert j} = \mathbf{W}_{ij}\textbf{u}_i
\end{equation}
where the $\mathbf{W}_{ij}\in \mathbb{R}^{d_u\times d_{\hat{u}}}$ is a weighted matrix that given an activation vector $\textbf{u}_i$ provides a prediction vector $\hat{\textbf{u}}_{i\vert j}$. Not all the capsules at layer $L$ are similarly affecting the capsules at layer $L+1$. The procedure identifies the coupling coefficients $c_{ij}$ that express the importance of capsule $i$ at lower layer for capsule $j$ at a higher layer. The $c_{ij}$ are computed by applying the soft-max function over $b_{ij}$:

\begin{equation}
    c_{ij} = \frac{exp(b_{ij})}{\sum_{k}exp(b_{ik})}
    \label{cij}
\end{equation}
where $b_{ij}$ is log probability of capsule $i$ being coupled with capsule $j$. The $b_{ij}$ variable is initialised at 0, then it is updated $\alpha$ times, at each iteration of the \myRbA{ }procedure.
Then, the input vector of capsule $j$ is computed as the weighted sum of the probability vectors at capsule $i$ multiplied by the coupling coefficient:

\begin{equation}
    \textbf{s}_j = \sum_{i}c_{ij}\hat{\textbf{u}}_{j\vert i}
    \label{sj}
\end{equation}
By having the input vector of capsule $j$ at higher layer, the clustering procedure is implemented by computing the agreement among capsules, as shown in Fig.~\ref{fig:RoutingBAgreement}. The output vectors of the capsules layer represent the probability of an object of being present in the given input or not. These vectors can exceed value one, depending on the output, so to make the output vector represents a probability, a non linear squashing function is used to restrict the vector length to one, where $\textbf{s}_j$ is input to capsule $j$ and $\textbf{v}_j$ is the output. 
\begin{equation}
    \textbf{v}_{j} = \frac{\Vert\textbf{s}_j\Vert^2}{1+\Vert\textbf{s}_j\Vert^2}\frac{\textbf{s}_j}{\Vert\textbf{s}_j\Vert}
    \label{vj}
\end{equation}
Finally $b_{ij}$ are updated by computing the inner product of $\textbf{v}_j$ and $\hat{\textbf{u}}_{j\vert i}$. If two vectors agree, the product would be larger leading to longer vector length. We summarise the \myRbA{ }in Algorithm~\ref{RBA}.
\begin{algorithm}
\caption{\myRbA{ }algorithm~\cite{sabour2017dynamic}}\label{alg:RoutingBAgreement}
\label{RBA}
\begin{algorithmic}
\Require $\textbf{u}_i \in \mathbb{R}^{d_u}$
\Ensure $\hat{\textbf{u}}_{i\vert j} = \mathbf{W}_{ij}\textbf{u}_i$
\For{all capsule $i \in [0,\beta]$ in layer L}
\For{all capsule $j \in [0,\beta]$ in layer L+1}
\State $b_{ij} \gets 0$
\For{$\alpha$ iteration of routing}
\State $c_{ij} = \frac{exp(b_{ij})}{\sum_{k}exp(b_{ik})}$ (Eq.~\ref{cij})\\
\State $\textbf{s}_j = \sum_{i}c_{ij}\hat{\textbf{u}}_{j\vert i}$ (Eq.~\ref{sj})\\
\State $\textbf{v}_{j} = \frac{\Vert\textbf{s}_j\Vert^2}{1+\Vert\textbf{s}_j\Vert^2}\frac{\textbf{s}_j}{\Vert\textbf{s}_j\Vert}$ (Eq.~\ref{vj})\\
\State $b_{ij} \gets b_{ij} + \textbf{v}_{j}\hat{\textbf{u}}_{j\vert i}$\\
\EndFor
\EndFor
\EndFor
\end{algorithmic}
\end{algorithm}
\subsection{Clustering of representative learning}
\label{C_Q}
Clustering implies partitioning data into meaningful groups such that items are similar within each group and dissimilar across different groups. We observe that capsules with \myRbA{ }implement a clusterization procedure. In Eq.~\ref{cij}, the coupling coefficients $c_{ij}$ is large for $(i, j)$ if capsule $i\in L$ is meaningful for capsule $j\in L+1$, and small otherwise. Such that $\hat{\textbf{u}}_{i\vert j}$ is similar to others $\hat{\textbf{u}}_{k\vert j}$, with $k\neq i$, if both depict meaningful information for capsule $j\in L+1$.   

In Fig~\ref{fig:RoutingBAgreement}, $\hat{\textbf{u}}_{i\vert j}$, depicted by blue circles, denote the agreements at layer $L$. These agreements are empowered by $c_{ij}$ in the clusterization phase, shown with the "Clustering" arrow, to obtain $\textbf{v}_j$. By squashing $\textbf{s}_{j}$, we interpret $\textbf{v}_{j}$ as the probability that capsule $i$ is grouped to cluster $j$. The $\textbf{v}_{j}$ for all the $j$ in $L$, represents the capsules vectors for the input. 
We think that the clusterization obtained by the capsules layer is a tradeoff between the random latent variables learned in \myVAE{ }and the discrete latent representation obtained with \myVQVAE{ }. In contrast with \myVQVAE, in \myCVGAN{ }we have the differentiability of the entire structure with the \myRbA{ }procedure and similar to the \myVQVAE{ }, we introduce the clusterization property of the latent variables which reduces the latent space generated by the model. 

\subsection{\myCVGAN{ }architecture}

\begin{figure}
  \centering
  \includegraphics[width=.8\linewidth]{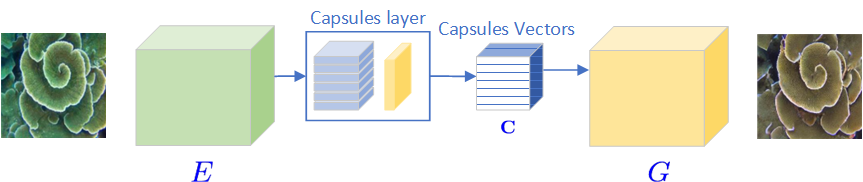}
  \caption{The proposed \myCVGAN: \myVAE{ }with newly proposed a latent space obtained by the capsules layer which implement the clustering of features.}
  \label{fig:CVVAE}
\end{figure}

In Fig.~\ref{fig:CVVAE}, we present the new model \myCVGAN. It is a convolutional model consisting of an encoder $E$, a new latent space based on capsules layer $CL$, and a decoder $G$, such that they learn to represent images with latent variables from a learned representation made by capsules vectors. We follow the implementation proposed in \myVQGAN{ }for the $E$ and $G$ networks. The $E$ network consists of six blocks with $\{ResNet-Attention-Convolution\}$. $E$ receives as input an image $\textbf{Y}\in \mathbb{R}^{c_{input}\times h_{input}\times w_{input}}$ and extracts an hidden/latent representation $\textbf{X}\in \mathbb{R}^{c_{z}\times h_{z}\times w_{z}}$ defined in the latent space $z$. $z(\textbf{X})$ is the input of $CL$. 
\begin{figure}
  \centering
  \includegraphics[width=.8\linewidth]{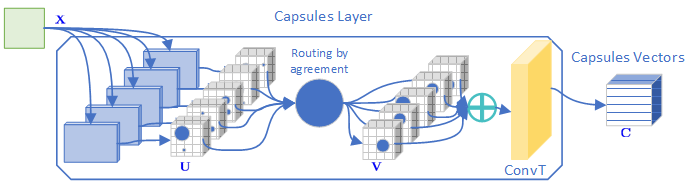}
  \caption{Proposed $CL$ of \myCVGAN: it consists of a capsules layer and a convolution transpose layer. The capsules extract $\textbf{U}$ features which are clusterised by the \myRbA{ }procedure, to obtain $\textbf{V}$. We aggregate the matrices and upsample by the convolutional transposed layer.}
  \label{fig:CV}
\end{figure}
$CL$ has $\beta$ capsules implemented with convolutional layers that compute $\beta$ different views, $\textbf{U}$, of the input $\textbf{X}$, Fig.~\ref{fig:CV}. The $\textbf{U} \in \mathbb{R}^{\beta\times d_u\times h\times w}$ consists of the activity vectors, each of which with $d_u$ digits. They are collectively produced by neurons in capsules at layer $L$. Each view in $\textbf{U}$ consists of $h\times w$ vectors. With their length and orientation, these vectors represent the presence and properties of entities identified in the image. Following the \myRbA{ }procedure described in Sec.~\ref{C_L}, the capsules layer is able to cluster the extracted $\textbf{U}$ features, and to remove the unclustered ones. In particular, the \myRbA{ }maintains the features depicted as present by the agreement among the capsules. The $CL$ provides $\textbf{V}\in \mathbb{R}^{\beta\times d_a\times h\times w}$, where $d_a$ is the number of digits of each vector obtained. We aggregate the features in $\textbf{V}$ by computing the $L_{1}norm$ of each vector, flattening $\textbf{V}$ to $\mathbb{R}^{\beta\times h\times w}$. Finally, a convolutional layer maps the flattened $\textbf{V}$ into $\textbf{C}\in \mathbb{R}^{c_C\times h_C \times w_C}$ that is the input of the decoder network. 

The decoder network $G$ is composed by six blocks, each of which is composed of $\{ResNet-Attention-Upsampling\}$. This network reconstructs the image from $\textbf{C}$ to generate an output image $\hat{\textbf{Y}} \in \mathbb{R}^{c_{input}\times h_{input}\times w_{input}}$ where the noise due to the underwater situation is removed.  

\subsection{Loss Function}
The proposed \myCVGAN{ }is trained to compress the essential features of the image in the capsules vectors at the latent space. We want that these features are the information about the subjects present in the image and that the model leaves behind the noise information. With this intent, we consider paired training samples for high-quality enhancement and we train the pretrained \myCVGAN{ }to reconstruct the input image as it is. The model is then fine-tuned for the denoising task. We observe that the pretraining of the model let the model build progressively the ability of reconstruction and denoise of the images . 

\subsubsection{Adversarial loss}
We train the model with the adversarial training procedure with a patch-based discriminator $D$~\cite{isola2017image} that aims to differentiate between real and reconstructed images.
The generator $G$ is implemented with the proposed model and it is conditioned on an underwater image \textbf{I} given as input to the model. We want the model to produce an image to try and fool $D$. Meanwhile, $D$ is trained to distinguish between the original not distorted images and the images generated by $G$. We follow the original formulation of $\mathcal{L}_{GAN}$ proposed in~\cite{esser2021taming}. 
\begin{equation}
\mathcal{L}_{GAN}({E, G, CL}, D) = \lambda[log D(\textbf{Y}) + log(1 - D(\hat{\textbf{Y}}))]  
\end{equation}
The $\hat{\textbf{Y}}$ denotes the output obtained by $G$ while $\textbf{Y}$ is the ground-truth image. The contribution of the $\mathcal{L}_{GAN}$ is controlled by factor $\lambda$ that is an adaptive weight computed according to:
\begin{equation}
    \lambda = \frac{\nabla_{G_{L}}[\mathcal{L}_{rec}]}{\nabla_{G_{L}}[\mathcal{L}_{GAN}] + \delta}
\label{lambda}
\end{equation}
where $\mathcal{L}_{rec}=\Vert\textbf{Y}-\hat{\textbf{Y}}\Vert^2 $ is the perceptual reconstruction loss~\cite{zhang2018unreasonable}, $\nabla_{G_{L}}[\cdot]$ is the gradient of its input at the last layer L of the decoder, and $\delta$ is used for numerical stability~\cite{esser2021taming}.

\subsubsection{Gradient difference loss}
We observe that generative models produce blurry images while reconstructing the underwater image. For this reason, in fine-tuning we add a sharpener factor to the loss function. This sharpener factor consists of directly penalise the differences of image gradient predictions in the generative loss function. We apply the Gradient Difference Loss (GDL) function~\cite{mathieu2015deep}. This function between the ground truth image $\textbf{Y}$ , and the prediction $\hat{\textbf{Y}}$ is given by:
\begin{equation}
   \mathcal{L}_{GDL}(\textbf{Y},\hat{\textbf{Y}}) =  \sum_{ij} \Vert\textbf{Y}_{ij}-\textbf{Y}_{i-1j}\vert - \vert\hat{\textbf{Y}}_{ij}-\hat{\textbf{Y}}_{i-1j}\Vert^\gamma + \Vert\textbf{Y}_{ij-1}-\textbf{Y}_{ij}\vert - \vert\hat{\textbf{Y}}_{ij-1}-\hat{\textbf{Y}}_{ij}\Vert^\gamma 
\label{GDL}
\end{equation}
where $\gamma$ is an integer greater or equal to 1, and $\vert\cdot\vert$ denotes the absolute value function. The GDL penalises gradient differences between the reconstructed and the real output. In this work, the image gradient is obtained by considering the neighbour pixel intensities differences, rather than adopting a more sophisticated norm on a larger neighbourhood, to keep the training time low.
\subsubsection{Combined loss}
In our experiments, we apply the $\mathcal{L}_{\myCVGAN}$ for the network \myCVGAN{ }, during pretraining. It is defined as:
\begin{equation}
\mathcal{L}_{\myCVGAN} =  \mathcal{L}_{rec} + \mathcal{L}_{GAN}({E, G, CL}, D)
\label{soegan_1}
\end{equation}
The function applied for \myUWCVGAN{ } obtained by "Underwater images training phase" considers the combination with the $\mathcal{L}_{GDL}$ to promote a sharper reconstruction. The contribution of the GDL function is considered in the ablation study at Sec.~\ref{Ablation}. In this phase we define $\mathcal{L}_{\myUWCVGAN}$ as:
\begin{equation}
\mathcal{L}_{\myUWCVGAN} =  \mathcal{L}_{rec} + \mathcal{L}_{GAN}({E, G, CL}, D) + \mathcal{L}_{GDL}
\label{soegan_2}
\end{equation}

\section{Results}\label{sec2}
\subsection{Dataset}
\label{dataset}
We consider \myCVGAN{ }pretrained on ImageNet dataset and we fine-tune the model for the underwater images task. Following~\cite{esser2021taming}, we consider the ImageNet dataset~\cite{russakovsky2015imagenet} and we pretrain \myCVGAN{ }for image reconstruction on the training split suggested for the ImageNet dataset consisting of 1.3M images (with no labels). The input images are resized to $256\times 256$. We identify the model at this stage with the name \myCVGAN.
\paragraph{Training phase: }
The \myCVGAN{ }pretrained on ImageNet is fine-tuned for the image denoising task on the training split suggested for UFO120 dataset. The UFO120 dataset~\cite{islam2020simultaneous} comprises 1500 samples shot underwater with no labels. Each sample consists of a noisy and denoised images pair. The noisy image is shot underwater and it shows water distortion, while the denoised image does not have the water distortion. We identify the \myCVGAN{ }fine-tuned for underwater images with the name \myUWCVGAN.
\paragraph{Validation phase: }
We evaluate the model over four benchmarks: Enhancing Underwater Visual Perception (EUVP), Heron Island Coral Reef Dataset (HICRD), Underwater Image Enhancement Benchmark (UIEB), and Underwater Image Super-Resolution (USR248). The dataset EUVP~\cite{islam2020fast} consists of  separate sets of paired and unpaired image samples of poor and good perceptual quality to facilitate supervised training of underwater image enhancement models. We take into consideration the validation split of paired images, which consists of 1970 images. The HICRD~\cite{CSIRO} is a set of images shot to the coral reef at deep sea with 300 pairs of images. The UIEB~\cite{peng2021ushape} includes 877 images, which involve richer underwater scenes (lighting conditions, water types, and target categories) and better visual quality reference images than the existing ones. Finally, the USR248~\cite{islam2020simultaneous} contains 248 samples of underwater images. The latter two datasets consist only of the underwater images with not denoised pairs.

\subsection{Implementation details}
The input the \myUWCVGAN is $\textbf{Y}\in \mathbb{R}^{3\times 256\times 256}$. The network $E$ outputs $z(\textbf{X})\in\mathbb(R)^{256\times 16\times 16}$, which is the input of $CL$. This consists of $\beta=32$ capsules and outputs $\textbf{U} \in \mathbb{R}^{32\times 16 \times9 \times9}$ through convolutional layers. The $[32\times 9\times 9]$ tensors, which are extracted by the capsules, describe the different points of view of the capsules and each vector has $d_u=16$ digits. In \myRbA{ }, we set $\alpha = 3$ for the $c_{i,j}$ loop update. The output of \myRbA{ }is $\textbf{V}\in \mathbb{R}^{32\times 64\times 9\times 9}$, where each vector obtained by clusterisation has $d_a=64$ digits. The transposed convolution layer in the capsules layer outputs $\textbf{C} \in \mathbb{R}^{256\times 16\times 16}$. 

Finally, $G$ decodes $\textbf{C}$ to generate the output matrix $\hat{\textbf{Y}}\in \mathbb{R}^{3\times 256 \times 256}$. The \myUWCVGAN{ }models $ X=3\times 256 \times 256$ images by compressing them to $z = 256\times 16\times 16$ latent space via the capsules layer $p(\textbf{X}\vert z)$. In particular, the compression factor obtained with the latent representation $z$ is:
\begin{equation}
    \frac{3\times 256\times 256}{256\times 16\times 16}=3
\end{equation}

We consider the batch size set to $6$. The \myCVGAN{ }is pretrained to reconstruct the input image $\textbf{Y}$ while we train the \myUWCVGAN{ }to reconstruct the denoised input denoted with $\textbf{Y}_{d}\in \mathbb{R}^{3\times 256\times 256}$. 
For the experimentation, the proposed models, \myCVGAN{ }and \myUWCVGAN, is trained respectively for $25$ and $500$ epochs.
Finally, we set $\delta = 10^{-6}$ in Eq.~\ref{lambda}, as suggested in~\cite{esser2021taming}, and $\gamma = 1$ in Eq.~\ref{GDL}.

\subsection{Metrics}
The \myUWCVGAN{ }model is evaluated qualitatively, by considering the visual presentation of the reconstructed and enhanced images, and quantitatively, by comparing our metrics results with the SOTA. In particular, we consider the Inception Score (IS)~\cite{salimans2016improved}, the Peak Signal-to-Noise Ratio (PSNR)~\cite{hore2010image}, the Underwater Color Image Quality Evaluation Metric (UCIQE)~\cite{7300447}, and the Underwater Image Quality Measure (UIQM)~\cite{7305804}. PSNR, IS metrics analyse the reconstruction quality obtained by the model, while UCIQE and UIQM are specific for the underwater images enhancement. The IS evaluates the quality of the image obtained by generative models by applying an Inception v3 Network pretrained on ImageNet. This quality is calculated as a statistic of the network’s outputs when applied to generated images. We apply IS according to project~\cite{obukhov2020torchfidelity}.
The PSNR evaluates the quality between the original and a reconstructed image. The higher the PSNR, the better the quality of the reconstructed image. Even if \myUWCVGAN{ }is a generative model, we consider the PSNR a valid metric to evaluate the reconstruction ability of the model.
The UCIQE is a linear combination of chroma, saturation, and contrast. It is proposed to quantify the non-uniform colour cast, blurring, and low-contrast that characterised underwater engineering and monitoring images. Finally, the UIQM comprises three underwater image attribute measures: the underwater image colorfulness measure (UICM), the underwater image sharpness measure (UISM), and the underwater image contrast measure (UIConM).
Each attribute is used to assess each singular aspect of the underwater image degradation. Therefore, the UIQM is the combination of the three, as $ UIQM = c_{1}UICM * c_{2}UISM * c_{3}UIConM$ where the colorfulness, sharpness, and contrast measures are linearly combined. We set the three parameters c1, c2, and c3 to 0.0282, 0.2953, and 3.5753 according to the paper~\cite{7305804}.

% IMAGE RESULTS
\begin{figure}
  \centering
  \includegraphics[width=0.8\linewidth]{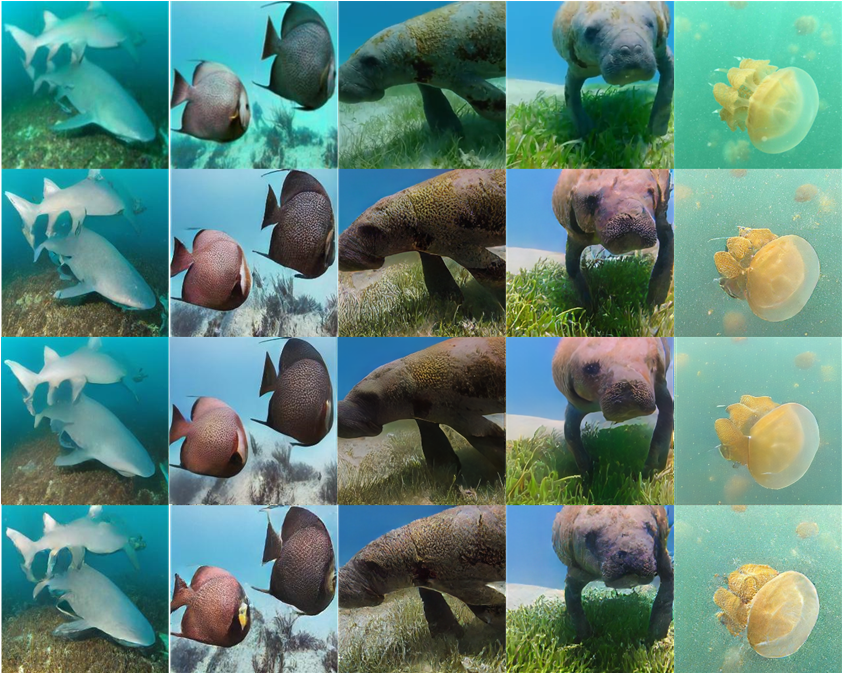}
  \caption{Ablation analysis: First row shows the input images. In the second row, the results of \myUWCVGAN{ }w/o $\mathcal{L}_{GDL}$, and in the third row of \myUWCVGAN{ }with $\mathcal{L}_{GDL}$. Last row, we present results obtained with the \myVQGAN{ }model, that is the direct competitor with the \myUWCVGAN.The samples are from the EUVP test split dataset and UIEB dataset.}
  \label{fig:ablationResults}
\end{figure}
\begin{table}[h]
\begin{center}
\caption{Ablation Study Metrics: We summarise the results obtained with the UCIQE, UIQM, PSNR, and IS metrics. The table compares different implementations of the \myUWCVGAN{ }trained with/without $\mathcal{L}_{GDL}$, and the \myVQGAN.}
\label{tab1}%
\begin{tabular}{@{}llcccc@{}}
\toprule
dataset& model & UCIQE$\uparrow$& UIQM$\uparrow$ & PSNR$\uparrow$&IS$\uparrow$ \\
\midrule
\multirow{3}{*}{EUVP} &UW-CVGAN & \textbf{ 6.47} & \textbf{2.91}& \textbf{29.11} &  4.39 ± 0.5  \\
&UW-CVGAN w/o GDL & 6.32  & 2.50& 28.93 & \textbf{5.02} ± 0.1 \\
&VQGAN  & 6.31& 2.39 & 28.90 & 4.82 ± 0.2 \\
\midrule
\multirow{3}{*}{HICRD}& UW-CVGAN &  \textbf{3.59} & \textbf{2.15}& \textbf{28.11}  &  1.87 ± 0.1 \\
&UW-CVGAN w/o GDL  & 3.40  & 2.05& 28.09 & \textbf{1.96} ± 0.1\\
&VQGAN & 3.32 & 1.95 & 28.10& 1.93 ± 0.1 \\
\midrule
\multirow{3}{*}{UIEB}& UW-CVGAN  &  \textbf{5.46} & \textbf{3.41}& - &  \textbf{3.73} ± 0.3 \\
&UW-CVGAN w/o GDL& 5.26  & 2.64& - &3.60 ± 0.4   \\
&VQGAN & 5.34 & 2.38 & - & 3.38 ± 0.4 \\
\midrule
\multirow{3}{*}{USR248}& UW-CVGAN &  6.46 &\textbf{2.43}& - &  4.72 ± 0.6 \\
&UW-CVGAN w/o GDL &6.21 & 2.03 & - & \textbf{5.03} ± 0.8\\
&VQGAN  & \textbf{6.53} & 1.90 & - & 4.93 ± 0.5  \\

\botrule
\end{tabular}
\end{center}
\end{table}

\subsection{Ablation study}
\label{Ablation}
The ablation study aims to reveal the effect of two main characteristics of the proposed model: 1) the impact of each component applied in the loss function; 2) the comparison between the usage of the Capsules vectors in \myUWCVGAN{ } and the discrete quantization, \myVQGAN{}. For this second point, the two models are trained under the same condition and with the same dataset. All the models presented in the ablation study are pretrained on ImageNet and fine-tuned on UFO120, as specified in Sec.~\ref{dataset}. Fig.~\ref{fig:ablationResults} shows the reconstruction of images from EUVP and UIEB datasets. 

In the samples shown in Fig.~\ref{fig:ablationResults}, the first row presents the input image from the original dataset. We compare results obtained with and without the application of the $\mathcal{L}_{GDL}$ component. The second row shows the reconstruction obtained with \myUWCVGAN{ }where the $\mathcal{L}_{GDL}$ component is not added to the $\mathcal{L}_{\myUWCVGAN}$. In this case, the $\mathcal{L}_{\myUWCVGAN}$ corresponds to Eq.~\ref{soegan_1}. We compare these results with the third row images, where the  $\mathcal{L}_{GDL}$ is present in  $\mathcal{L}_{\myUWCVGAN}$ defined by Eq.~\ref{soegan_2}. We observe that the quality of the images obtained with $\mathcal{L}_{\myUWCVGAN}=$Eq.~\ref{soegan_2} is higher, and the contours appear well defined. This observation is supported by the quantitative results presented in Tab.~\ref{tab1}, which summarises the metric results obtained with the four benchmark datasets proposed for validation. The results obtained with \myUWCVGAN{ }are higher in almost all considered metrics compared to the results provided by \myUWCVGAN{ }w/o GDL. Based on this observation, we demonstrate that the application of the $\mathcal{L}_{GDL}$ provides fundamental information for the reconstruction of the image reaching high-quality results. The PSNR, metric is not available for UIEB and USR248 datasets because the ground truth is not available at SOTA.

We now compare the results obtained with \myUWCVGAN{ }(third row) and results with \myVQGAN{ }(fourth row). In Fig.~\ref{fig:ablationResults} and Tab.~\ref{tab1}, we observe that the application of capsules layer in the \myUWCVGAN, provides a visible improvement in the quality of the reconstruction compared with the reconstruction obtained with \myVQGAN{ }. In the third row (from the top), the images appear to have higher quality with well-defined edges of the entities represented compared to the images at the fourth row (from the top). This observation is visible also in metrics (Tab.~\ref{tab1}), where \myUWCVGAN{ } achieves better results in almost all the datasets considered.

\subsection{Results Comparison}
\label{resultsComp}
We consider the Deep Sesr~\cite{islam2020simultaneous}, Funie gan~\cite{islam2020fast}, Ugan and Ugan-p~\cite{fabbri2018enhancing} works to provide a qualitative and quantitative comparison evaluation. The evaluation presented is performed on the EUVP dataset, commonly used for evaluation by all the works considered for comparison.
\begin{figure}
  \centering
  \includegraphics[width=\linewidth]{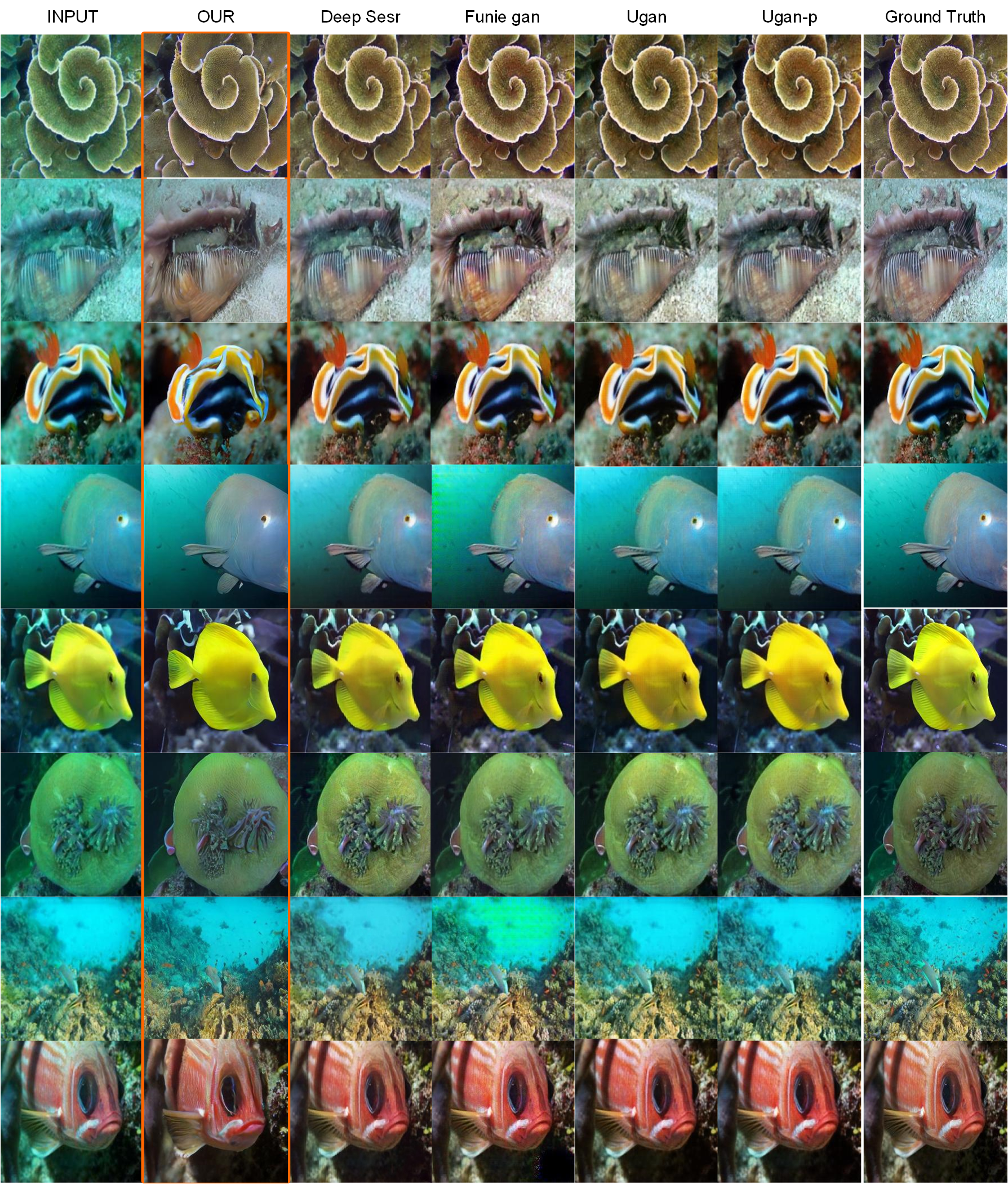}
  \caption{Comparison of the reconstructed images: The first column shows the input images provided to the models for the denoising task. We compare our results with EUVP dataset (second column from the left, in the orange rectangle), with results obtained with models at the state of the art Deep Sesr~\cite{islam2020simultaneous}, Funie gan~\cite{islam2020fast}, Ugan and Ugan-p~\cite{fabbri2018enhancing}. Last column is the ground truth of the denoised images.} 
  \label{fig:ResultsComparison}
\end{figure}
\begin{figure}
  \centering
  \includegraphics[width=\linewidth]{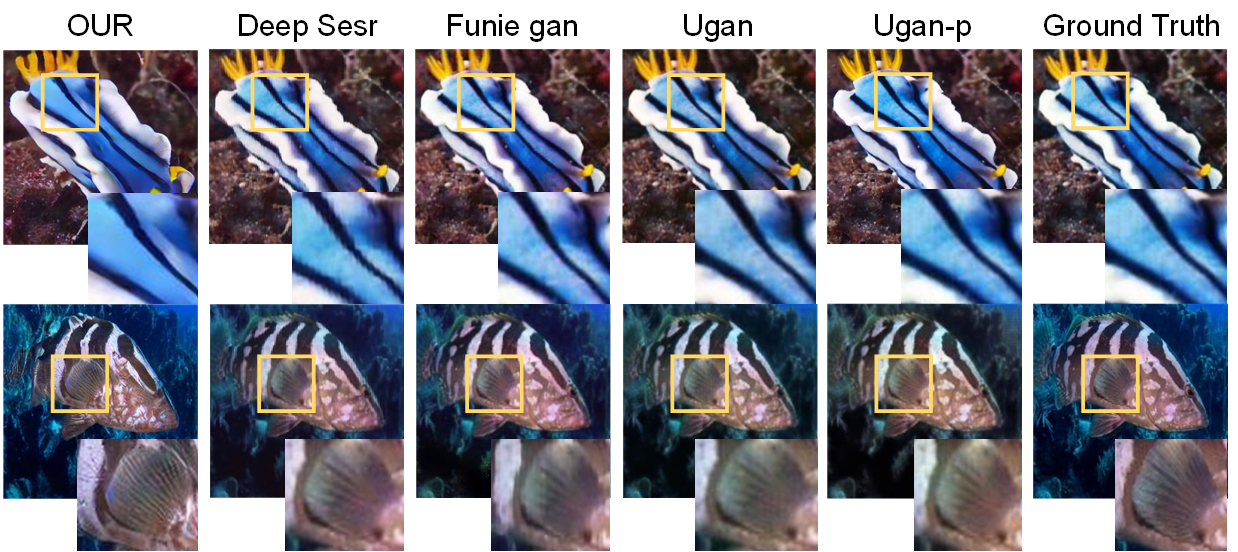}
  \caption{Focus of reconstruction quality of images: The first column shows the denoising and reconstruction of images from EUVP dataset with proposed \myUWCVGAN{ }(second column from the left). We compare our model with results obtained with models Deep Sesr~\cite{islam2020simultaneous}, Funie gan~\cite{islam2020fast}, Ugan and Ugan-p~\cite{fabbri2018enhancing}. Last column is the ground truth of the denoised images.} 
  \label{fig:ResultsComparisonFocus}
\end{figure}
In Fig.~\ref{fig:ResultsComparison}, we qualitatively compare the reconstructed images provided by \myUWCVGAN{ }(second column) and by the others works(third-sixth columns). We observe that the images provided by \myUWCVGAN{ }have neat images and overall better quality than the SOTA. In all the images, the subjects are reconstructed with a bright colourisation and a wealth of details. 

To support this observation, in Fig.~\ref{fig:ResultsComparisonFocus}, we show two samples from EUVP dataset focusing on details in the images. The yellow square identifies the area of the images that is the object of the zoom. In the first row, we present a blue and white sea slug body. In the first column, the reconstructed body of the animal has a high quality definition of details and the colours appear to be smooth and bright. On the contrary, the reconstructions obtained with SOTA methods have low in quality.  In the second row, we zoom on the detail of the fin of the fish. We observe results similar to the slug: the reconstruction obtained with our model outperforms other methods for image quality and the definition of details. It is worth noting that these results demonstrate how the \myUWCVGAN{ }is able to reconstruct the enhanced input image by decoding only the latent variables.
\begin{table}[h]
\begin{center}
\caption{Results Comparison Metrics: We summarise the results obtained on UCIQE, UIQM, PSNR, and IS metrics. We present the results obtained with \myUWCVGAN{ }, Deep Sesr~\cite{islam2020simultaneous}, Funie gan~\cite{islam2020fast}, Ugan and Ugan-p~\cite{fabbri2018enhancing} computed over the EuVP dataset. The last two columns present the memory space required to store the underwater image or its latent features.}
\label{tab2}%
\begin{tabular}{@{}lcccc|cc@{}}
\toprule
model& UCIQE$\uparrow$& UIQM$\uparrow$ & PSNR$\uparrow$  &IS$\uparrow$ & Params &Bytes\\
\midrule
\textbf{\myUWCVGAN{ }(OUR)}&\textbf{ 6.47} & \textbf{2.91 }& \textbf{29.11} & \textbf{4.39 ± 0.5} & $256\times 16\times16$ &\textbf{262KB}\\
Deep Sesr& 5.86 & 2.69& 29.39  & 4.99 ± 0.7 &$116\times72\times32$&1.1MB\\
Funie gan & 6.66 & 2.07 & 29.41 & 5.25 ± 0.8 & $256\times256\times3$&786KB\\
Ugan  & 6.00 & 2.15 & 29.34& 5.24 ± 0.7 &$256\times256\times3$&786KB\\
Ugan-p & 6.30 & 2.12& 29.31 & 5.08 ± 0.7 &$256\times256\times3$&786KB\\
\botrule
\end{tabular}
\end{center}
\end{table}
Tab.~\ref{tab2} shows results obtain on the EUVP dataset. We compare \myUWCVGAN{ }with the models at SOTA over the metrics considered for evaluation. We observe that our proposed method achieves results that are in line with SOTA in all the metrics. The last two columns in Tab.~\ref{tab2} denote the number of parameters per image/latent representation and the byte required for storing them. We observe that \myUWCVGAN{ }can store the only latent representation which consists of less parameters than the original input image. In Deep Sesr~\cite{islam2020simultaneous}, the dimension of the encoded image requires $1.1MB$ required while with \myUWCVGAN{ }it is less than $300KB$. In contrast with Funie gan~\cite{islam2020fast}, Ugan and Ugan-p~\cite{fabbri2018enhancing}, $E$ and $G$ in \myUWCVGAN{ }are not entangled with skip connection resulting in the possibility of storing only the encoded image ($256\times 16\times16$). 

We demonstrate, that \myUWCVGAN{ }is able to reach metrics results on par with the other methods while requiring less memory storage per image by storing only the encoded image when the encoder is embedded onto the collector.
\begin{figure}
  \centering
  \includegraphics[width=\linewidth]{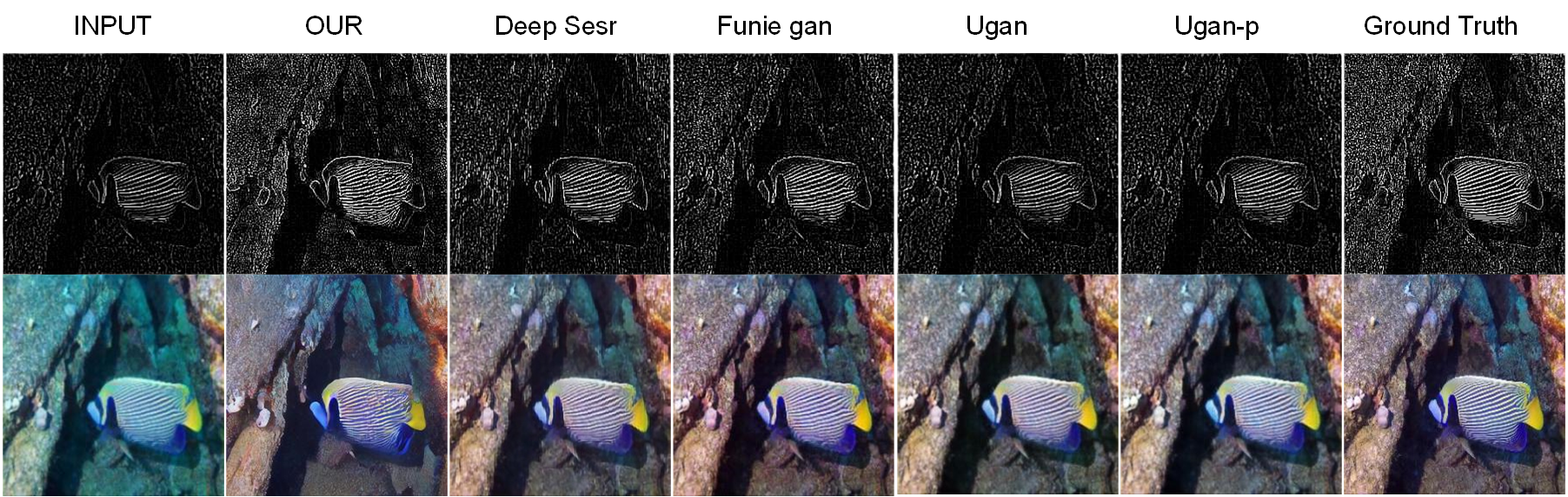}
  \caption{Focus of reconstruction quality of images: running the Canny Edge Detector on sample images. The first row shows the edges identified by the canny edge detector over the reconstructed images obtained with the underwater input, \myUWCVGAN, and models at the state of the art Deep Sesr~\cite{islam2020simultaneous}, Funie gan~\cite{islam2020fast}, Ugan and Ugan-p~\cite{fabbri2018enhancing}. Last column is the ground truth denoised images.} 
  \label{fig:ResultsCannyEdgeFocus}
\end{figure}
 \begin{figure}
  \centering
  \includegraphics[width=\linewidth]{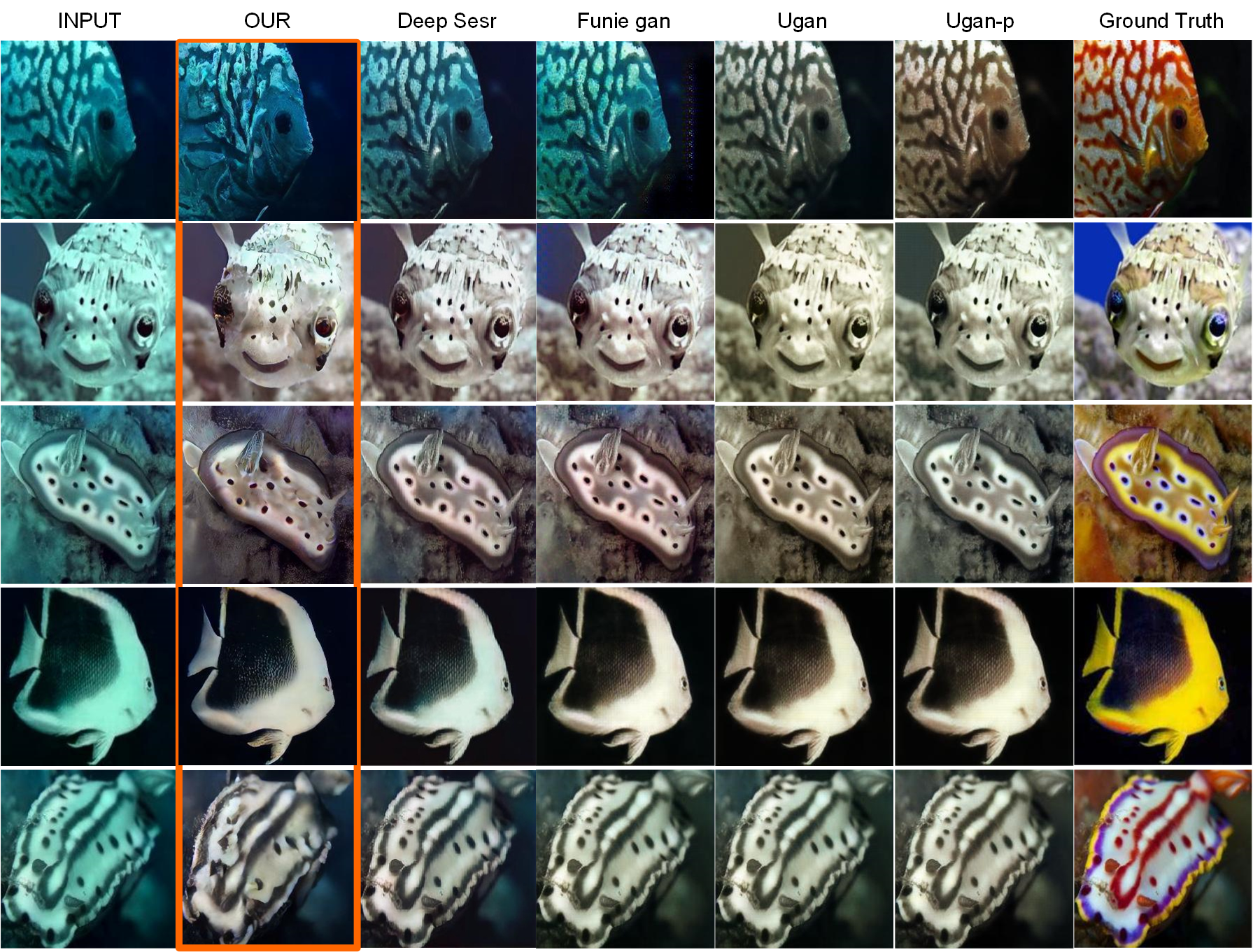}
  \caption{A sample of failure in images reconstruction with \myUWCVGAN{ }compared with SOTA: the first column is the input images provided to the models from EUVP dataset. We compare our results, with results obtained with models at the state of the art Deep Sesr~\cite{islam2020simultaneous}, Funie gan~\cite{islam2020fast}, Ugan and Ugan-p~\cite{fabbri2018enhancing}. Last column is the ground truth denoised images.} 
  \label{fig:ResultsComparisonfail}
\end{figure}

Finally, we use the Canny edge detector~\cite{canny1986computational}, which provides a colour agnostic evaluation of the images compared to ground truth. In Fig.~\ref{fig:ResultsCannyEdgeFocus}, we show the edges detected images obtained with \myUWCVGAN{ }compared to the ones obtained with the other methods. We observe that the edges obtained with our methods appear to be less noisy then the other methods and visibly close to the edges in the ground truth. In Tab.~\ref{tab3}, we consider the Euclidean distance computed between the edges detected in the input image, the edges detected in images reconstructed with each model considered for comparison. We observe that the edges obtained with \myUWCVGAN{ }has a lower distance from edges of the ground truth compared to the other methods. This demonstrates that the enhanced images provided by \myUWCVGAN{ }outperform them in the definition of details.

\begin{table}[h]
\begin{center}
\caption{Results Comparison Euclidean distance of Canny Edge: We summarise the Euclidean distance computed between the edges detected in original input, the edges detected in images obtained with \myUWCVGAN{ }compared with results obtained with \myUWCVGAN{ }, Deep Sesr~\cite{islam2020simultaneous}, Funie gan~\cite{islam2020fast}, Ugan and Ugan-p~\cite{fabbri2018enhancing} computed over the EuVP dataset. The results presented are to be considered multiplied to $10E5$.}
\label{tab3}%
\begin{tabular}{@{}lccccc@{}}
\toprule
 & \myUWCVGAN{ }(OUR) & Deep Sesr & Funie gan & Ugan & Ugan-p\\
L2Norm &\textbf{344.36}&349.89&358.89&352.58&353.35\\
\botrule
\end{tabular}
\end{center}
\end{table}
\section{Discussion}\label{sec12}
The results presented in Sec.~\ref{resultsComp} demonstrate the ability of \myUWCVGAN{ }in image enhancement of an underwater image. In contrast with Funie gan~\cite{islam2020fast}, Ugan, and Ugan-p~\cite{fabbri2018enhancing}, \myUWCVGAN{ }reconstructs the images only from the latent space keeping the encoder and decoder networks separated. This characteristic allow to use the encoder to compress the image while collecting. In contrast with Deep Sesr~\cite{islam2020simultaneous}, \myUWCVGAN{ }compresses the input image in the latent representation by a factor of 3, while Deep Sesr compress the features by a factor of 1.5. 
The \myUWCVGAN{ }faces a challenging reconstruction of underwater images learning to extract the latent representation of the image which is used to provide the enhanced image.

Finally, we think it is interesting to observe the model's failures. In Fig.~\ref{fig:ResultsComparisonfail}, we present samples that are wrongly reconstructed. The reconstructed images appear to be desaturated, and the colourisation is brownish or greenish. This result is obtained with \myUWCVGAN{ }as with works at SOTA and highlights an interesting limit of these generative methods in colours reconstruction in underwater images. This problem is evident mainly with the input images where the red components are suppressed due to the light absorption in the water and opens up to the need of further study with such types of models.
\section{Conclusion}
In this paper, we demonstrate the application of the capsules layer as a quantization layer for \myVAE{ }by proposing \myCVGAN. The new architecture is a valid tradeoff between a fully differentiable model and discrete latent features, with the compromise of dynamic learned of quantization. We propose this model for underwater image enhancement, and we train the model to reconstruct while enhancing the distorted images. We obtain results on par with SOTA on UCIQE, UIQM, PSNR, and IS metrics. Moreover, our model is an AutoEncoder where the encoder network is completely independent of the decoder. This brings up two observations: the model is able to extract representative features that are left behind the noise of the water with the only features extracted by the encoder. That proves that the encoder can be used as compressing algorithm to reduce the input image to its fundamental features obtaining $3\times$ of compression factor. The model is the first attempt at a generative variational AutoEncoder where the capsules layer is used to perform the quantization of the latent space, and the first attempt of such a model for the underwater image enhancement. The results obtained show up high quality in reconstruction and bright and saturated colours. We finally analyse failures in image reconstruction that open up to other new challenges.

\bibliography{sn-bibliography}% common bib file
% %% if required, the content of .bbl file can be included here once bbl is generated
% %%\input sn-article.bbl

% %% Default %%
% %%\input sn-sample-bib.tex%

\end{document}